\documentclass[11pt]{article}

\usepackage[preprint]{acl}

\usepackage{times}
\usepackage{latexsym}
\usepackage{tabularx}
\usepackage[T1]{fontenc}

\usepackage[utf8]{inputenc}

\usepackage{microtype}

\usepackage{inconsolata}
\usepackage{tikz}
\usetikzlibrary{arrows.meta}

\usepackage{graphicx}
\usepackage{booktabs}
\usepackage{enumitem}
\usepackage{multirow}

\usepackage{tikz}

\usepackage{booktabs}
\usepackage{tabularx}
\usepackage{ragged2e}
\usepackage{xcolor}
\usepackage{pifont}

\definecolor{okgreen}{HTML}{1B7F3B}
\definecolor{badred}{HTML}{B02418}
\definecolor{cgrey}{HTML}{444444}

\newcolumntype{L}{>{\RaggedRight\arraybackslash}X}
\newcommand{\gloss}[1]{\par\addvspace{2pt}\textit{\textcolor{cgrey}{#1}}}
\newcommand{\note}[1]{\par\addvspace{3pt}{\footnotesize\color{cgrey}#1}}
\newcommand{\ok}{\textcolor{okgreen}{\ding{51}}}
\newcommand{\bad}{\textcolor{badred}{\ding{55}}}
\newcommand{\err}[1]{\textcolor{badred}{#1}}
\newcommand{\fix}[1]{\textcolor{okgreen}{#1}}

\newcommand{\ourSet}{SiLTT}

\usetikzlibrary{positioning}

\title{Precision over Scale: A Polish–Silesian Benchmark and a Translation System Outperforming Open-Source and Commercial Models}

\author{
 \textbf{Grzegorz Kulik\textsuperscript{1}} 
  \quad
 \textbf{Mikołaj Pokrywka\textsuperscript{1}}
  \quad
 \textbf{Adam Jatowt\textsuperscript{2}}
  \quad
 \textbf{Wojciech Kusa\textsuperscript{1}}
  \vspace{0.6em}
\\
 \textsuperscript{1}NASK National Research Institute, Poland
 \quad
 \textsuperscript{2}University of Innsbruck, Austria
\vspace{0.5em}
\\
 \small{
   \textbf{Correspondence:} \href{mailto:email@domain}{name.surname@nask.pl}
 }
}

\begin{document}
\maketitle
\begin{abstract}


Dialectal machine translation (MT) remains challenging due to limited data and strong linguistic variation not captured by standard benchmarks, which often assume standardized and well-edited text.
We study Polish--Silesian MT using neural and rule-based systems, evaluating on \ourSet{} -- a new Pol--Szl testset, alongside established BOUQuET and FLORES benchmarks. 
Results show our rule-based system is consistently strongest on \ourSet{} and BOUQuET datasets and that TranslateGemma fine-tuned on a curated dataset improves over strong neural baselines but does not surpass the rule-based system in dialectal settings. 
We release \ourSet{} and our best neural model to support further research.\footnote{\url{https://huggingface.co/datasets/NASK-PIB/SiLTT}}\textsuperscript{,}\footnote{\url{https://huggingface.co/NASK-PIB/translategemma-4b-it-pol-szl-qlora}}
\end{abstract}

\section{Introduction}

Silesian (ISO 639-3: \textit{szl}) is a West Slavic regional language of Poland with limited NLP resources and an ongoing debate about its linguistic status.
According to the National Population and Housing Census 2021, 467,145 residents of Poland declared Silesian as a language used in home contacts \cite{gus2021}.
Despite growing interest in low-resource machine translation, Silesian remains largely absent from dedicated MT benchmarks and system comparisons.

A key challenge in developing Silesian language technologies is not only the scarcity of parallel data, but also its uneven quality. While Silesian text is present in online sources, much of it is inconsistent in orthography, heavily influenced by Polish, or shaped by ad-hoc writing conventions rather than stable linguistic norms. As a result, large-scale corpora derived from web data or generic multilingual resources often mix heterogeneous and noisy forms of Silesian, limiting their usefulness for training robust translation systems.

Existing multilingual benchmarks such as FLORES-101/200~\cite{guzman2019flores,goyal2022flores} include Silesian as part of broad evaluation coverage. However, they primarily reflect formal, edited text and do not capture the variability and informal usage found in real-world Polish–Silesian communication. This mismatch becomes more pronounced when models are trained on large but noisy corpora (e.g., OPUS and other web-mined datasets), where Silesian-like data often lacks linguistic consistency and contains substantial cross-lingual contamination.



In this paper, we investigate Polish–Silesian machine translation under realistic, data-constrained conditions. Rather than relying on large, noisy corpora, we construct a compact but curated parallel dataset of approximately 22k sentence pairs designed to better reflect usable Silesian. We compare neural and rule-based with commercial and other state of the art multilingual translation systems.

Our results show that systems trained on large, heterogeneous corpora do not necessarily generalize well to cleaner dialectal evaluation settings. In contrast, our rule-based system remains consistently strong, while neural models trained on the curated dataset outperform strong baselines but still fall short of rule-based performance in dialect-sensitive evaluation.



Our contributions are the following:
\begin{itemize}[noitemsep,leftmargin=1em]
    \item SiLTT: a new Polish--Silesian translation evaluation dataset;
    \item a systematic comparison of neural and rule-based MT systems across three benchmarks;
    \item a publicly released evaluation dataset and fine-tuned neural model.
\end{itemize}

\section{Related Work}

Machine translation for low-resource and dialectal language varieties remains challenging due to limited parallel data and high linguistic variation~\cite{haddow2022survey,ranathunga2023neural}. 
Recent multilingual systems such as mT5~\cite{xue2021mt5}, NLLB~\cite{nllbteam2022languageleftbehindscaling}, and OPUS-MT~\cite{tiedemann2024democratizing} showed that transfer learning and massively multilingual training can improve translation quality for under-resourced languages. 
Several studies further demonstrated that transfer between related languages is particularly effective in low-resource and zero-shot MT settings~\cite{zoph2016transfer,dabre2017empirical,lakew2019adapting}. 

Silesian is included in two multilingual evaluation benchmarks: FLORES-101/200~\cite{guzman2019flores,goyal2022flores} and BOUQuET~\cite{theomnilingualmtteam2025bouquetdatasetbenchmarkopen}, enabling evaluation of multilingual MT systems on this language. 
However, FLORES has seen extensive and repeated use in multilingual MT evaluation, which can lead to benchmark saturation effects.

The closest work to ours is~\citet{nlp:raslan23conf/pp145-156}, who studied zero-shot Silesian--English translation using mT5 and ByT5 models fine-tuned on related Slavic languages. They found that cross-lingual transfer from related languages can improve performance, though the effect depends on model architecture and language similarity. Their evaluation relied on FLORES due to the lack of dedicated resources for Silesian.



\section{Language Background}

Silesian’s orthography was standardized only in 2009 with the ślabikŏrz system, following centuries of ad hoc adaptations of Polish orthography or phonetic scripts ~\cite{bandtkie,steuer}. Its status as a distinct language or Polish dialect remains debated: some classify it as an Abstand language, isolated from Polish and neighboring dialects ~\cite{hentschel_2018}, while others argue for mutual intelligibility and shared features with Polish dialects ~\cite{dubisz_2024,Siuciak_2025}. Lack of official recognition limits its use in education, and most speakers rely on Polish orthography, reinforced by the absence of default Silesian keyboard support. 
Silesian exhibits variation across orthography, regional morphology, and lexical borrowings from neighboring languages (Polish, Czech, German). These properties make direct sentence-level translation challenging for purely data-driven systems, especially in low-resource regimes
For more details, see the Appendix~\ref{app:silesian_language}.

\section{System Description}

\subsection{Rule-Based System}
\label{sec:rule-based-system}

We implement a rule-based Polish--Silesian translator built on the \texttt{Apertium} platform~\cite{forcada2011apertium,khanna2021recent}. The system follows a shallow-transfer architecture and includes: (1) finite-state transducers for lexical transformations, (2) morphological analysis and generation, (3) Hidden Markov Models for POS tagging and ambiguity resolution,
and (4) handcrafted transfer rules encoding systematic correspondences.


Our system contains 31,757 lexical translation pairs and 2,349 grammatical rules derived from dictionaries \cite{przywara,olesch,zareba,pluta,czastkaszymon}, as well as theses and articles on Silesian vocabulary (e.g., \citet{kaisig,nikel,koszyk,stawoj}).
We have included a detailed illustrated system pipeline in Appendix~\ref{app:pipeline}.


\begin{figure}[htbp]
  \centering
\resizebox{0.95\columnwidth}{!}{
\begin{tikzpicture}[
  box/.style={
    draw=black!80, rectangle, rounded corners=2pt,
    minimum width=1.55cm, minimum height=0.72cm,
    align=center, font=\scriptsize\sffamily,
    fill=gray!10, line width=0.5pt
  },
  io/.style={draw=none, align=center, font=\scriptsize\itshape},
  pipe/.style={
    -{Stealth[length=4pt, width=3pt]},
    line width=0.75pt, color=black!70
  }
]
\node[box] (def)    at (0.00, 0)    {de-\\[-1pt]formatter};
\node[box] (morph1) at (1.90, 0)    {morph.\\[-1pt]anal.};
\node[box] (pos)    at (3.80, 0)    {PoS\\[-1pt]tagger};
\node[box] (struct) at (3.80,-1.25) {struct.\\[-1pt]transf.};
\node[box] (lex)    at (1.90,-1.25) {lex.\\[-1pt]transfer};
\node[box] (morph2) at (0.00,-1.25) {morph.\\[-1pt]gen.};
\node[box] (post)   at (0.00,-2.50) {post-\\[-1pt]generator};
\node[box] (ref)    at (1.90,-2.50) {re-\\[-1pt]formatter};
\node[io]  (sl)     at (0.00, 0.90) {Source Language text};
\node[io]  (tl)     at (4.10,-2.50) {Target Language text};
\draw[pipe] (sl) -- (def);
\draw[pipe] (def) -- (morph1);  \draw[pipe] (morph1) -- (pos);
\draw[pipe] (pos) -- (struct);
\draw[pipe] (struct) -- (lex);  \draw[pipe] (lex) -- (morph2);
\draw[pipe] (morph2) -- (post);
\draw[pipe] (post) -- (ref);
\draw[pipe] (ref) -- (tl);
\end{tikzpicture}
}
  \caption{The eight modules that build the assembly line of the shallow-transfer
machine translation system. 
}
  \label{fig:your-label}
\end{figure}












\subsection{Neural models}
To evaluate the performance of various neural machine translation architectures on the Silesian language, we selected a representative suite of models ranging from specialized multilingual systems to large-scale commercial language models (details in Appendix~\ref{sec:neural_models}):

\begin{itemize}[noitemsep,leftmargin=1em]
    \item \textbf{PLLuM-12B-nc-chat}: Polish general-purpose LLM~\cite{kocoń2025pllumfamilypolishlarge}
    \item \textbf{Bielik-PL-11B-v3.0-IT}: Polish-focused multilingual model with claimed Silesian support~\cite{ociepa2025bielik11bv3multilingual}
    \item \textbf{NLLB-200-3.3B, NLLB-54B MoE}: Silesian-capable MT models~\cite{nllbteam2022languageleftbehindscaling}
    \item \textbf{MADLAD-400-10B-MT}: multilingual MT model based on the T5 architecture~\cite{kudugunta2023madlad400}
    \item \textbf{GPT-5.4}: frontier zero-shot LLM~\cite{singh2026openaigpt5card}
    \item \textbf{Google Translate}: a production-grade MT system~\cite{wu2016googlesneuralmachinetranslation}
    \item \textbf{TranslateGemma-4B-IT (base, fine-tuned)}: instruction-tuned and custom finetuned MT variants~\cite{finkelstein2026translategemmatechnicalreport}
\end{itemize}


\section{Datasets}

\subsection{Training Data}
\label{sec:training_data}
Our Polish-Silesian training data consists of three complementary sources. In particular, we combine manually curated and post-edited data with automatically generated resources obtained via pivoting through English and backtranslation. All data was filtered using a maximum length ratio threshold of 1.3.

\begin{itemize}[noitemsep,leftmargin=1em]
    \item \textbf{22{,}783 sentences — Custom backtranslated dataset of Silesian sentences (CBT-22k):} We used articles provided to us by the websites Wachtyrz.eu and Slazag.pl as well as several older dialect texts, the largest being the collection gathered by \citet{malinowski}. The older texts were transliterated to the current alphabet. The Silesian sentences were translated into Polish using the rule-based system described in Section~\ref{sec:rule-based-system}, and subsequently post-edited by a professional translator to ensure high-quality references. \label{custom_train_data}

    \item \textbf{50{,}000 sentences — NLLB corpora (pivoting via English):} Sampled from the OPUS corpora~\cite{tiedemann-nygaard-2004-opus}, specifically the NLLB dataset~\cite{schwenk2020ccmatrixminingbillionshighquality,fan2020englishcentricmultilingualmachinetranslation}, which contains approximately 1.7 million English--Silesian pairs. The English side was translated into Polish using a quantized version of EuroLLM-9B-Instruct~\cite{martins2025eurollm9btechnicalreport} model.\footnote{\scriptsize \url{https://huggingface.co/stelterlab/EuroLLM-9B-Instruct-AWQ}} We decided to sample only 50,000 sentences to avoid overflowing our custom high-quality dataset (CBT-22k).

    \item \textbf{28{,}107 sentences — HPLT backtranslation} Extracted from the monolingual Silesian HPLT v3.0 corpus~\cite{oepen2026hplt30largescalemultilingual}. Silesian sentences were translated into Polish using the NLLB-3.3B model to obtain a back-translated dataset.
\end{itemize}

\subsection{Evaluation Data}

We use three evaluation datasets:


\textbf{BOUQuET}: \cite{theomnilingualmtteam2025bouquetdatasetbenchmarkopen} a multilingual, multicentric benchmark consisting of 854 Polish--Silesian sentence pairs translated by community contributors. We additionally conducted a manual evaluation with native Silesian and Polish speaker to verify overall quality of the translations.

\textbf{\ourSet{}} (\textbf{Si}lesian \textbf{L}iterary \textbf{T}ranslation \textbf{T}estset): 1,237 Polish--Silesian sentence pairs derived from human translations and automatically aligned using LaBSE~\cite{feng2022languageagnosticbertsentenceembedding}. The dataset was compiled from literary and cultural corpora, including ``\textit{Cinderella}'', ``\textit{Rapunzel}'', ``\textit{Janko the Musician}'' and ``\textit{A Christmas Carol}''.
%
%
We applied automatic alignment filtering based on cosine similarity of sentence embeddings, retaining only pairs with similarity above 0.6. Sentence pairs below this threshold were excluded from the final dataset. Dataset statistics are shown at Table~\ref{tab:stats}.
The resulting test set is released under the 
CC BY-NC 4.0
license.

\begin{table}[h]
\centering
\label{tab:testset_stats}
\resizebox{\columnwidth}{!}{
\begin{tabular}{lcc}
\hline
\textbf{Statistic} & \textbf{Polish} & \textbf{Silesian} \\
\hline
Avg sentence length (chars) & 93.1 & 79.4 \\
Avg sentence length (words) & 14.8 & 13.6 \\
Max sentence length (chars) & 542 & 445 \\
Max sentence length (words) & 82 & 73 \\
Vocabulary size       & 7673 & 5393 \\
\hline
\end{tabular}
}
\caption{Dataset statistics of \ourSet{}.}
\label{tab:stats}
\end{table}

\textbf{FLORES-200}: \cite{guzman2019flores} A multilingual evaluation benchmark consisting of 1{,}012 high-quality sentence pairs for each language pair.

\begin{table}[th]
\centering
\resizebox{\columnwidth}{!}{
\begin{tabular}{lrr|rr}
\toprule
\multirow{2}{*}{\textbf{Model}} & \multicolumn{2}{c|}{\textbf{\ourSet{}}} & \multicolumn{2}{c}{\textbf{BOUQuET}} \\
 & \textbf{BLEU} & \textbf{chrF} & \textbf{BLEU} & \textbf{chrF} \\
\midrule
PLLuM-12B-nc-chat & 1.7 & 22.3 & 3.4 & 28.4 \\
Bielik-PL-11B-v3.0-IT & 3.4 & 26.3 & 7.8 & 35.5 \\ \midrule
MADLAD-400-10B-MT & 2.1 & 21.3 & 7.5 & 32.0 \\ 
NLLB 3.3B & 3.8 & 26.8 & 12.2 & 39.2 \\
NLLB 54B MOE & 3.2 & 26.0 & 9.9 & 38.3 \\ 
TransGemma & 2.5 & 23.6 & 7.4 & 30.9 \\ \midrule
GPT-5.4 & 7.4 & 31.7 & 20.0 & 48.6 \\
Google Translate & 4.2 & 29.3 & 21.8 & 49.4 \\ \midrule
\small{\textit{ours}} \\
Rule-Based & \textbf{10.3} & \textbf{34.5} & \textbf{36.7} & \textbf{61.0} \\
TransGemma-FT (CBT-22k) & \underline{8.0} & \underline{31.9} & \underline{26.3} & \underline{52.8} \\
TransGemma-FT (Mixed) & 5.5 & 28.9 & 6.2 & 33.4 \\
\bottomrule
\end{tabular}
}
\caption{Comparison of POL$\rightarrow$SZL results on \ourSet{} and BOUQuET benchmarks. \textbf{Bold} indicates the best result, and \underline{underlining} indicates the second-best result.}
\label{tab:combined_results}
\end{table}

\section{Experiments}
\label{sec:experiments}
We primarily focus on the Polish-to-Silesian translation direction, as it represents the more challenging task. Compared to Silesian, Polish is a high-resource language, making text generation substantially easier for modern models.

\subsection{Fine-tuning the TranslateGemma Model}
We fine-tuned the 4-billion-parameter instruction-tuned version of TranslateGemma\footnote{\scriptsize \url{https://huggingface.co/google/translategemma-4b-it}} using QLoRA~\cite{dettmers2023qloraefficientfinetuningquantized}.
We performed two fine-tuning experiments with different training data configurations: (1) using only the custom dataset of Silesian sentences (CBT-22k), and (2) using all corpora described in Section~\ref{sec:training_data}. This setup was motivated by concerns, based on manual inspection, that the Silesian side of the NLLB- and HPLT-derived data may contain noise and low-quality texts. Due to the low-resource nature of Silesian, 
reliable automatic filtering methods are not available.
\subsection{Evaluation Metrics}
We use sacreBLEU~\cite{post2018clarityreportingbleuscores} to calculate the BLEU\footnote{\scriptsize BLEU signature: "nrefs:1|case:mixed|eff:no|tok:13a|smooth:exp|version:2.6.0}~\cite{papineni-etal-2002-bleu} and chrF
\footnote{\scriptsize chrF signature: "nrefs:1|case:mixed|eff:yes|nc:6|nw:0|space:no|version:2.6.0}~\cite{popovic-2015-chrf} scores. We do not use neural-based evaluation metrics such as COMET~\cite{rei-etal-2020-comet} or LLM-as-a-Judge approaches, as the low-resource setting may lead to unreliable and potentially misleading results in this context.

\begin{table}[h]
\centering
\small

\begin{tabular}{lrr}
\toprule
\textbf{Model} & \textbf{BLEU} & \textbf{chrF} \\
\midrule
PLLuM-12B-nc-chat & 0.1 & 12.1 \\
Bielik-PL-11B-v3.0-IT & 5.7 & 35.1 \\ \midrule
MADLAD-400-10B-MT & 9.8 & 37.2 \\
NLLB 3.3B & 7.4 & 34.8 \\
NLLB 54B MOE & 7.4 & 37.5 \\
TransGemma & 5.1 & 32.1 \\ \midrule
GPT-5.4 & 11.9 & 44.2 \\
Google Translate & \textbf{17.7} & \textbf{48.6} \\ \midrule
\small{\textit{ours}} \\
Rule-Based & \underline{14.9} & \underline{48.3} \\
TransGemma-FT (CBT-22k) & 12.9 & 45.8 \\
TransGemma-FT (Mixed) & 10.4 & 41.4 \\
\bottomrule
\end{tabular}
\caption{Results on the FLORES benchmark for POL$\rightarrow$SZL translation. Google Translate achieves the best performance overall, followed by the rule-based system. TranslateGemma fine-tuned on CBT-22k improves over the base model but remains below the strongest systems.}
\label{table:flores}
\end{table}

\begin{table*}[t]
\centering
\small
\setlength{\tabcolsep}{5pt}
\renewcommand{\arraystretch}{1.1}
\begin{tabularx}{\linewidth}{@{}>{\RaggedRight\arraybackslash}p{3.2cm} L L L@{}}
\toprule
 & \multicolumn{3}{c}{\textbf{System output}} \\
\cmidrule(l){2-4}
\textbf{Source (PL)} & \textbf{Rule-based} & \textbf{TransGemma-FT}\newline\textbf{(CBT-22k)} & \textbf{Google Translate} \\
\midrule

Chłopcy i dziewczynki byli bardzo szczęśliwi podczas zabawy.
\gloss{Boys and girls were very happy during play.}
& Synki i dziouszki byli fest szczyńśliwi w czasie graczki.
  \note{\ok~Correct.}
& \err{Chłopcy} i \err{dziouchy} byli bardzo szczyńśliwi podczas \err{zabawy}.
  \note{\bad~Polish retained: \emph{chłopcy}~$\rightarrow$~\fix{synki}, \emph{zabawy}~$\rightarrow$~\fix{graczki}. Wrong diminutive: \emph{dziouchy}~$\rightarrow$~\fix{dziouszki}.}
& \err{Syny} i \err{dziouchki} byli barzo szczyńśliwi w czasie \err{szpilu}.
  \note{\bad~Morphology: \emph{syny}~$\rightarrow$~\fix{synki}, \emph{dziouchki}~$\rightarrow$~\fix{dziouszki}. Lexical: \emph{szpilu} (`game')~$\rightarrow$~\fix{graczki}.}
\\
\midrule

Możesz też skontaktować się z nami przez telefon.
\gloss{Alternatively, you can reach out to us by phone.}
& Możesz tyż skōntaktować sie z nami bez telefōn.
  \note{\ok~Correct.}
& Możesz tyż skōntaktować sie z nami \err{przez} telefōn.
  \note{\bad~Polish retained: \emph{przez}~$\rightarrow$~\fix{bez}.}
& Możesz tyż sie z \err{niymi} \err{spokopić} bez telefōn.
  \note{\bad~Person: \emph{niymi} (`them')~$\rightarrow$~\fix{nami}. Lexical: \emph{spokopić} (`understand')~$\rightarrow$~\fix{skōntaktować}.}
\\
\midrule

I przyprowadzę dzieci.
\gloss{And I'll bring the kids.}
& I przikludza dzieci.
  \note{\ok~Correct.}
& I przikludza dzieci.
  \note{\ok~Correct.}
& A \err{ja} \err{prziwiozōm} dzieci.
  \note{\bad~Redundant Polish \emph{ja}~$\rightarrow$~\fix{jo}. Person and lexis: \emph{prziwiozōm} (`they will bring by car')~$\rightarrow$~\fix{przikludza}/\fix{prziwiyda}.}
\\
\midrule

Twój ojciec za bardzo się martwi.
\gloss{Your father worries too much.}
& Twōj fater za fest sie \err{umrzici}.
  \note{\bad~\emph{martwi} read as `the dead' rather than `worries'; the two are homographs in Polish.}
& Twōj ôjciec za bardzo sie \err{martwi}.
  \note{\bad~Polish retained: \emph{martwi}~$\rightarrow$~\fix{tropi}/\fix{staro}.}
& Twōj fater sie za moc tropi.
  \note{\ok~Correct.}
\\
\midrule
Rzeczy, które chciałbym wiedzieć wcześniej.
\gloss{Things I wish I had known earlier.}
& Rzeczy, co \err{ôna} chcioł bych wiedzieć wczaśnij.
  \note{\bad~Case of \emph{rzeczy} misread: the singular genitive and plural nominative are homographs in both Polish and Silesian, giving the wrong gender and case in \emph{ôna}~$\rightarrow$~\fix{je}.}
& Rzeczy, kere chcioł bych wiedzieć wczaśnij.
  \note{\ok~Correct.}
& Rzeczy, kere chciołbych \err{wczasnij}.\ \err{}
  \note{\bad~Misspelling: \emph{wczasnij}~$\rightarrow$~\fix{wczaśnij}.}
\\
\bottomrule
\end{tabularx}
\caption{Qualitative comparison on the BOUQuET dataset. Source sentences are given with English glosses. Within each system output, tokens marked in red are erroneous; the expected Silesian form is given in green in the accompanying note.}
\label{tab:qualitative_examples}
\end{table*}

\subsection{Results}
The evaluation results on \ourSet{}, the BOUQuET POL$\rightarrow$SZL task, and FLORES are presented in Tables~\ref{tab:combined_results} and~\ref{table:flores}, while results for the reverse direction (SZL$\rightarrow$POL) are provided in Appendix~\ref{app:expanded_results}.

The rule-based system achieves the best results on \ourSet{} and BOUQuET, maintaining a clear advantage on both benchmarks, while remaining competitive with the strongest systems on FLORES.
The base TranslateGemma model performs poorly on all evaluation sets, indicating limited zero-shot capability for this low-resource language pair. However, fine-tuning on the CBT-22k corpus yields substantial improvements, producing the strongest neural system and surpassing both GPT-5.4 and Google Translate. These results highlight the importance of high-quality parallel data for effective adaptation in low-resource MT.
In contrast, augmenting the training data with additional NLLB and HPLT corpora consistently degrades performance compared to the CBT-22k-only setup, likely due to noise.


The Polish models Bielik and PLLuM also achieve relatively weak results across all evaluation settings. Although Bielik was pretrained on Silesian text, its performance remains limited, only slightly better than that of the base TranslateGemma model. PLLuM performs poorly, failing to generate usable translations.
Overall, strong Polish language ability or incidental exposure to Silesian is insufficient for effective Polish–Silesian MT without dedicated bilingual supervision.

\subsubsection{Flores}
As shown in Table~\ref{table:flores}, Google Translate achieves the best performance on the FLORES benchmark. This unusually strong performance may indicate that the model was exposed to FLORES sentences during training, particularly given that the benchmark has been publicly available for a long time. However, this advantage does not transfer to the \ourSet{} and BOUQuET datasets.

\subsection{Qualitative Analysis}

We conduct an qualitative analysis of selected best models on the BOUQuET dataset (see Table~\ref{tab:qualitative_examples}). 
Key observation: explicit linguistic rules provide robustness in low-resource dialect translation.
The rule-based system provides the best results in the context of closely related languages. Its weaknesses usually have their origins in homographs. This could be remedied by creating more rules but we have not estimated neither the count of the remaining homographs, nor the time needed to create rules for them. Google Translate seems to create plausible but hallucinated phrases. Many mistakes made by that tool stem from its translating through English. The resulting text often has pronouns usually redundant in Slavic languages. At the same time the lack of person markers in English verbs leads to incorrect verb forms in output sentences. Fine-tuned TransGemma high-quality CBT-22k dataset provides promising results despite minor mistakes. The frequency of its hallucinations is low. Mistakes are usually made by keeping original words in the output text.

All systems lack the ability to rephrase sentences to a more probable word order and style, eg. "Rzeczy, kere chcioł bych wiedzieć wczaśnij" seen in the last example in the Table~\ref{tab:qualitative_examples} is a grammatically correct sentence, however "Rzeczy, co bych je chcioł wiedzieć wczaśnij" would be a more often used one in live speech.

\section{Conclusion}

We presented a new benchmark and evaluation of Polish--Silesian machine translation systems. Our findings suggest that rule-based systems remain highly competitive in dialectal and low-resource settings, particularly when evaluated on newer datasets.
We open-source the \ourSet{} testset and the best performing neural model.


\section*{Limitations}

This study is limited by the size and scope of the available evaluation and training data. Although the CBT-22k corpus is carefully curated, it remains relatively small compared to large-scale multilingual MT datasets, which may constrain the generalization of the neural models.

Our evaluation focuses on Polish--Silesian translation and three benchmark datasets; while these cover both curated and existing benchmark settings, they do not fully capture the breadth of real-world Silesian usage across domains, speakers, and registers. In particular, spoken language and highly informal communication styles are underrepresented.

The rule-based system relies on manually designed correspondences and linguistic assumptions, which may not scale to broader coverage or other dialectal language pairs without significant adaptation effort. While it performs strongly in the evaluated setting, its portability and maintenance cost remain open questions.

\bibliography{custom}

@inproceedings{guzman2019flores,
  title={The FLORES evaluation datasets for low-resource machine translation: Nepali--English and Sinhala--English},
  author={Guzm{\'a}n, Francisco and Chen, Peng-Jen and Ott, Myle and Pino, Juan and Lample, Guillaume and Koehn, Philipp and Chaudhary, Vishrav and Ranzato, Marc’Aurelio},
  booktitle={Proceedings of the 2019 conference on empirical methods in natural language processing and the 9th international joint conference on natural language processing (EMNLP-IJCNLP)},
  pages={6098--6111},
  year={2019}
}

@inproceedings{xue2021mt5,
  title={mT5: A massively multilingual pre-trained text-to-text transformer},
  author={Xue, Linting and Constant, Noah and Roberts, Adam and Kale, Mihir and Al-Rfou, Rami and Siddhant, Aditya and Barua, Aditya and Raffel, Colin},
  booktitle={Proceedings of the 2021 conference of the North American chapter of the association for computational linguistics: Human language technologies},
  pages={483--498},
  year={2021}
}

@article{goyal2022flores,
  title={The Flores-101 evaluation benchmark for low-resource and multilingual machine translation},
  author={Goyal, Naman and Gao, Cynthia and Chaudhary, Vishrav and Chen, Peng-Jen and Wenzek, Guillaume and Ju, Da and Krishnan, Sanjana and Ranzato, Marc’Aurelio and Guzm{\'a}n, Francisco and Fan, Angela},
  journal={Transactions of the Association for Computational Linguistics},
  volume={10},
  pages={522--538},
  year={2022},
  publisher={MIT Press One Broadway, 12th Floor, Cambridge, Massachusetts 02142, USA~…}
}

@inproceedings{lakew2019adapting,
  title={Adapting multilingual neural machine translation to unseen languages},
  author={Lakew, Surafel M and Karakanta, Alina and Federico, Marcello and Negri, Matteo and Turchi, Marco},
  booktitle={Proceedings of the 16th International Conference on Spoken Language Translation},
  year={2019}
}

@inproceedings{zoph2016transfer,
  title={Transfer learning for low-resource neural machine translation},
  author={Zoph, Barret and Yuret, Deniz and May, Jonathan and Knight, Kevin},
  booktitle={Proceedings of the 2016 conference on empirical methods in natural language processing},
  pages={1568--1575},
  year={2016}
}

@inproceedings{dabre2017empirical,
  title={An empirical study of language relatedness for transfer learning in neural machine translation},
  author={Dabre, Raj and Nakagawa, Tetsuji and Kazawa, Hideto},
  booktitle={Proceedings of the 31st Pacific Asia conference on language, information and computation},
  pages={282--286},
  year={2017}
}

@article{ranathunga2023neural,
  title={Neural machine translation for low-resource languages: A survey},
  author={Ranathunga, Surangika and Lee, En-Shiun Annie and Prifti Skenduli, Marjana and Shekhar, Ravi and Alam, Mehreen and Kaur, Rishemjit},
  journal={ACM Computing Surveys},
  volume={55},
  number={11},
  pages={1--37},
  year={2023},
  publisher={ACM New York, NY}
}

@article{haddow2022survey,
  title={Survey of low-resource machine translation},
  author={Haddow, Barry and Bawden, Rachel and Miceli-Barone, Antonio Valerio and Helcl, Jind{\v{r}}ich and Birch, Alexandra},
  journal={Computational Linguistics},
  volume={48},
  number={3},
  pages={673--732},
  year={2022}
}

@article{tiedemann2024democratizing,
  title={Democratizing neural machine translation with OPUS-MT},
  author={Tiedemann, J{\"o}rg and Aulamo, Mikko and Bakshandaeva, Daria and Boggia, Michele and Gr{\"o}nroos, Stig-Arne and Nieminen, Tommi and Raganato, Alessandro and Scherrer, Yves and V{\'a}zquez, Ra{\'u}l and Virpioja, Sami},
  journal={Language Resources and Evaluation},
  volume={58},
  number={2},
  pages={713--755},
  year={2024},
  publisher={Springer}
}

@inproceedings{nlp:raslan23conf/pp145-156,
    title={{Fine-Grained Language Relatedness for Zero-Shot Silesian-English Translation}},
    booktitle={Proceedings of Recent Advances in Slavonic Natural Language Processing, RASLAN 2023},
    author={Edoardo Signoroni},
    year={2023},
    pages={145--156},
    bibsource = {RASLAN OC, http://raslan2023.nlp-consulting.net/, paper ID 4}
}

@article{forcada2011apertium,
  title={Apertium: a free/open-source platform for rule-based machine translation},
  author={Forcada, Mikel L and Ginest{\'\i}-Rosell, Mireia and Nordfalk, Jacob and O’Regan, Jim and Ortiz-Rojas, Sergio and P{\'e}rez-Ortiz, Juan Antonio and S{\'a}nchez-Mart{\'\i}nez, Felipe and Ram{\'\i}rez-S{\'a}nchez, Gema and Tyers, Francis M},
  journal={Machine translation},
  volume={25},
  number={2},
  pages={127--144},
  year={2011},
  publisher={Springer}
}

@article{khanna2021recent,
  title={Recent advances in Apertium, a free/open-source rule-based machine translation platform for low-resource languages},
  author={Khanna, Tanmai and Washington, Jonathan N and Tyers, Francis M and Bayatl{\i}, Sevilay and Swanson, Daniel G and Pirinen, Tommi A and Tang, Irene and Alos i Font, Hector},
  journal={Machine Translation},
  volume={35},
  number={4},
  pages={475--502},
  year={2021},
  publisher={Springer}
}

@misc{nllbteam2022languageleftbehindscaling,
      title={No Language Left Behind: Scaling Human-Centered Machine Translation}, 
      author={NLLB Team and Marta R. Costa-jussà and James Cross and Onur Çelebi and Maha Elbayad and Kenneth Heafield and Kevin Heffernan and Elahe Kalbassi and Janice Lam and Daniel Licht and Jean Maillard and Anna Sun and Skyler Wang and Guillaume Wenzek and Al Youngblood and Bapi Akula and Loic Barrault and Gabriel Mejia Gonzalez and Prangthip Hansanti and John Hoffman and Semarley Jarrett and Kaushik Ram Sadagopan and Dirk Rowe and Shannon Spruit and Chau Tran and Pierre Andrews and Necip Fazil Ayan and Shruti Bhosale and Sergey Edunov and Angela Fan and Cynthia Gao and Vedanuj Goswami and Francisco Guzmán and Philipp Koehn and Alexandre Mourachko and Christophe Ropers and Safiyyah Saleem and Holger Schwenk and Jeff Wang},
      year={2022},
      eprint={2207.04672},
      archivePrefix={arXiv},
      primaryClass={cs.CL},
      url={https://arxiv.org/abs/2207.04672}, 
}

@misc{singh2026openaigpt5card,
      title={OpenAI GPT-5 System Card}, 
      author={Aaditya Singh and Adam Fry and Adam Perelman and Adam Tart and Adi Ganesh and Ahmed El-Kishky and Aidan McLaughlin and Aiden Low and AJ Ostrow and Akhila Ananthram and Akshay Nathan and Alan Luo and Alec Helyar and Aleksander Madry and Aleksandr Efremov and Aleksandra Spyra and Alex Baker-Whitcomb and Alex Beutel and Alex Karpenko and Alex Makelov and Alex Neitz and Alex Wei and Alexandra Barr and Alexandre Kirchmeyer and Alexey Ivanov and Alexi Christakis and Alistair Gillespie and Allison Tam and Ally Bennett and Alvin Wan and Alyssa Huang and Amy McDonald Sandjideh and Amy Yang and Ananya Kumar and Andre Saraiva and Andrea Vallone and Andrei Gheorghe and Andres Garcia Garcia and Andrew Braunstein and Andrew Liu and Andrew Schmidt and Andrey Mereskin and Andrey Mishchenko and Andy Applebaum and Andy Rogerson and Ann Rajan and Annie Wei and Anoop Kotha and Anubha Srivastava and Anushree Agrawal and Arun Vijayvergiya and Ashley Tyra and Ashvin Nair and Avi Nayak and Ben Eggers and Bessie Ji and Beth Hoover and Bill Chen and Blair Chen and Boaz Barak and Borys Minaiev and Botao Hao and Bowen Baker and Brad Lightcap and Brandon McKinzie and Brandon Wang and Brendan Quinn and Brian Fioca and Brian Hsu and Brian Yang and Brian Yu and Brian Zhang and Brittany Brenner and Callie Riggins Zetino and Cameron Raymond and Camillo Lugaresi and Carolina Paz and Cary Hudson and Cedric Whitney and Chak Li and Charles Chen and Charlotte Cole and Chelsea Voss and Chen Ding and Chen Shen and Chengdu Huang and Chris Colby and Chris Hallacy and Chris Koch and Chris Lu and Christina Kaplan and Christina Kim and CJ Minott-Henriques and Cliff Frey and Cody Yu and Coley Czarnecki and Colin Reid and Colin Wei and Cory Decareaux and Cristina Scheau and Cyril Zhang and Cyrus Forbes and Da Tang and Dakota Goldberg and Dan Roberts and Dana Palmie and Daniel Kappler and Daniel Levine and Daniel Wright and Dave Leo and David Lin and David Robinson and Declan Grabb and Derek Chen and Derek Lim and Derek Salama and Dibya Bhattacharjee and Dimitris Tsipras and Dinghua Li and Dingli Yu and DJ Strouse and Drew Williams and Dylan Hunn and Ed Bayes and Edwin Arbus and Ekin Akyurek and Elaine Ya Le and Elana Widmann and Eli Yani and Elizabeth Proehl and Enis Sert and Enoch Cheung and Eri Schwartz and Eric Han and Eric Jiang and Eric Mitchell and Eric Sigler and Eric Wallace and Erik Ritter and Erin Kavanaugh and Evan Mays and Evgenii Nikishin and Fangyuan Li and Felipe Petroski Such and Filipe de Avila Belbute Peres and Filippo Raso and Florent Bekerman and Foivos Tsimpourlas and Fotis Chantzis and Francis Song and Francis Zhang and Gaby Raila and Garrett McGrath and Gary Briggs and Gary Yang and Giambattista Parascandolo and Gildas Chabot and Grace Kim and Grace Zhao and Gregory Valiant and Guillaume Leclerc and Hadi Salman and Hanson Wang and Hao Sheng and Haoming Jiang and Haoyu Wang and Haozhun Jin and Harshit Sikchi and Heather Schmidt and Henry Aspegren and Honglin Chen and Huida Qiu and Hunter Lightman and Ian Covert and Ian Kivlichan and Ian Silber and Ian Sohl and Ibrahim Hammoud and Ignasi Clavera and Ikai Lan and Ilge Akkaya and Ilya Kostrikov and Irina Kofman and Isak Etinger and Ishaan Singal and Jackie Hehir and Jacob Huh and Jacqueline Pan and Jake Wilczynski and Jakub Pachocki and James Lee and James Quinn and Jamie Kiros and Janvi Kalra and Jasmyn Samaroo and Jason Wang and Jason Wolfe and Jay Chen and Jay Wang and Jean Harb and Jeffrey Han and Jeffrey Wang and Jennifer Zhao and Jeremy Chen and Jerene Yang and Jerry Tworek and Jesse Chand and Jessica Landon and Jessica Liang and Ji Lin and Jiancheng Liu and Jianfeng Wang and Jie Tang and Jihan Yin and Joanne Jang and Joel Morris and Joey Flynn and Johannes Ferstad and Johannes Heidecke and John Fishbein and John Hallman and Jonah Grant and Jonathan Chien and Jonathan Gordon and Jongsoo Park and Jordan Liss and Jos Kraaijeveld and Joseph Guay and Joseph Mo and Josh Lawson and Josh McGrath and Joshua Vendrow and Joy Jiao and Julian Lee and Julie Steele and Julie Wang and Junhua Mao and Kai Chen and Kai Hayashi and Kai Xiao and Kamyar Salahi and Kan Wu and Karan Sekhri and Karan Sharma and Karan Singhal and Karen Li and Kenny Nguyen and Keren Gu-Lemberg and Kevin King and Kevin Liu and Kevin Stone and Kevin Yu and Kristen Ying and Kristian Georgiev and Kristie Lim and Kushal Tirumala and Kyle Miller and Lama Ahmad and Larry Lv and Laura Clare and Laurance Fauconnet and Lauren Itow and Lauren Yang and Laurentia Romaniuk and Leah Anise and Lee Byron and Leher Pathak and Leon Maksin and Leyan Lo and Leyton Ho and Li Jing and Liang Wu and Liang Xiong and Lien Mamitsuka and Lin Yang and Lindsay McCallum and Lindsey Held and Liz Bourgeois and Logan Engstrom and Lorenz Kuhn and Louis Feuvrier and Lu Zhang and Lucas Switzer and Lukas Kondraciuk and Lukasz Kaiser and Manas Joglekar and Mandeep Singh and Mandip Shah and Manuka Stratta and Marcus Williams and Mark Chen and Mark Sun and Marselus Cayton and Martin Li and Marvin Zhang and Marwan Aljubeh and Matt Nichols and Matthew Haines and Max Schwarzer and Mayank Gupta and Meghan Shah and Melody Y. Guan and Melody Huang and Meng Dong and Mengqing Wang and Mia Glaese and Micah Carroll and Michael Lampe and Michael Malek and Michael Sharman and Michael Zhang and Michele Wang and Michelle Pokrass and Mihai Florian and Mikhail Pavlov and Miles Wang and Ming Chen and Mingxuan Wang and Minnia Feng and Mo Bavarian and Molly Lin and Moose Abdool and Mostafa Rohaninejad and Nacho Soto and Natalie Staudacher and Natan LaFontaine and Nathan Marwell and Nelson Liu and Nick Preston and Nick Turley and Nicklas Ansman and Nicole Blades and Nikil Pancha and Nikita Mikhaylin and Niko Felix and Nikunj Handa and Nishant Rai and Nitish Keskar and Noam Brown and Ofir Nachum and Oleg Boiko and Oleg Murk and Olivia Watkins and Oona Gleeson and Pamela Mishkin and Patryk Lesiewicz and Paul Baltescu and Pavel Belov and Peter Zhokhov and Philip Pronin and Phillip Guo and Phoebe Thacker and Qi Liu and Qiming Yuan and Qinghua Liu and Rachel Dias and Rachel Puckett and Rahul Arora and Ravi Teja Mullapudi and Raz Gaon and Reah Miyara and Rennie Song and Rishabh Aggarwal and RJ Marsan and Robel Yemiru and Robert Xiong and Rohan Kshirsagar and Rohan Nuttall and Roman Tsiupa and Ronen Eldan and Rose Wang and Roshan James and Roy Ziv and Rui Shu and Ruslan Nigmatullin and Saachi Jain and Saam Talaie and Sam Altman and Sam Arnesen and Sam Toizer and Sam Toyer and Samuel Miserendino and Sandhini Agarwal and Sarah Yoo and Savannah Heon and Scott Ethersmith and Sean Grove and Sean Taylor and Sebastien Bubeck and Sever Banesiu and Shaokyi Amdo and Shengjia Zhao and Sherwin Wu and Shibani Santurkar and Shiyu Zhao and Shraman Ray Chaudhuri and Shreyas Krishnaswamy and Shuaiqi and Xia and Shuyang Cheng and Shyamal Anadkat and Simón Posada Fishman and Simon Tobin and Siyuan Fu and Somay Jain and Song Mei and Sonya Egoian and Spencer Kim and Spug Golden and SQ Mah and Steph Lin and Stephen Imm and Steve Sharpe and Steve Yadlowsky and Sulman Choudhry and Sungwon Eum and Suvansh Sanjeev and Tabarak Khan and Tal Stramer and Tao Wang and Tao Xin and Tarun Gogineni and Taya Christianson and Ted Sanders and Tejal Patwardhan and Thomas Degry and Thomas Shadwell and Tianfu Fu and Tianshi Gao and Timur Garipov and Tina Sriskandarajah and Toki Sherbakov and Tomek Korbak and Tomer Kaftan and Tomo Hiratsuka and Tongzhou Wang and Tony Song and Tony Zhao and Troy Peterson and Val Kharitonov and Victoria Chernova and Vineet Kosaraju and Vishal Kuo and Vitchyr Pong and Vivek Verma and Vlad Petrov and Wanning Jiang and Weixing Zhang and Wenda Zhou and Wenlei Xie and Wenting Zhan and Wes McCabe and Will DePue and Will Ellsworth and Wulfie Bain and Wyatt Thompson and Xiangning Chen and Xiangyu Qi and Xin Xiang and Xinwei Shi and Yann Dubois and Yaodong Yu and Yara Khakbaz and Yifan Wu and Yilei Qian and Yin Tat Lee and Yinbo Chen and Yizhen Zhang and Yizhong Xiong and Yonglong Tian and Young Cha and Yu Bai and Yu Yang and Yuan Yuan and Yuanzhi Li and Yufeng Zhang and Yuguang Yang and Yujia Jin and Yun Jiang and Yunyun Wang and Yushi Wang and Yutian Liu and Zach Stubenvoll and Zehao Dou and Zheng Wu and Zhigang Wang},
      year={2026},
      eprint={2601.03267},
      archivePrefix={arXiv},
      primaryClass={cs.CL},
      url={https://arxiv.org/abs/2601.03267}, 
}

@misc{wu2016googlesneuralmachinetranslation,
      title={Google's Neural Machine Translation System: Bridging the Gap between Human and Machine Translation}, 
      author={Yonghui Wu and Mike Schuster and Zhifeng Chen and Quoc V. Le and Mohammad Norouzi and Wolfgang Macherey and Maxim Krikun and Yuan Cao and Qin Gao and Klaus Macherey and Jeff Klingner and Apurva Shah and Melvin Johnson and Xiaobing Liu and Łukasz Kaiser and Stephan Gouws and Yoshikiyo Kato and Taku Kudo and Hideto Kazawa and Keith Stevens and George Kurian and Nishant Patil and Wei Wang and Cliff Young and Jason Smith and Jason Riesa and Alex Rudnick and Oriol Vinyals and Greg Corrado and Macduff Hughes and Jeffrey Dean},
      year={2016},
      eprint={1609.08144},
      archivePrefix={arXiv},
      primaryClass={cs.CL},
      url={https://arxiv.org/abs/1609.08144}, 
}

@misc{finkelstein2026translategemmatechnicalreport,
      title={TranslateGemma Technical Report}, 
      author={Mara Finkelstein and Isaac Caswell and Tobias Domhan and Jan-Thorsten Peter and Juraj Juraska and Parker Riley and Daniel Deutsch and Geza Kovacs and Cole Dilanni and Colin Cherry and Eleftheria Briakou and Elizabeth Nielsen and Jiaming Luo and Kat Black and Ryan Mullins and Sweta Agrawal and Wenda Xu and Erin Kats and Stephane Jaskiewicz and Markus Freitag and David Vilar},
      year={2026},
      eprint={2601.09012},
      archivePrefix={arXiv},
      primaryClass={cs.CL},
      url={https://arxiv.org/abs/2601.09012}, 
}

@inproceedings{tiedemann-nygaard-2004-opus,
    title = "The {OPUS} Corpus - Parallel and Free: \url{http://logos.uio.no/opus}",
    author = {Tiedemann, J{\"o}rg  and
      Nygaard, Lars},
    editor = "Lino, Maria Teresa  and
      Xavier, Maria Francisca  and
      Ferreira, F{\'a}tima  and
      Costa, Rute  and
      Silva, Raquel",
    booktitle = "Proceedings of the Fourth International Conference on Language Resources and Evaluation ({LREC}{'}04)",
    month = may,
    year = "2004",
    address = "Lisbon, Portugal",
    publisher = "European Language Resources Association (ELRA)",
    url = "https://aclanthology.org/L04-1174/"
}

@misc{schwenk2020ccmatrixminingbillionshighquality,
      title={CCMatrix: Mining Billions of High-Quality Parallel Sentences on the WEB}, 
      author={Holger Schwenk and Guillaume Wenzek and Sergey Edunov and Edouard Grave and Armand Joulin},
      year={2020},
      eprint={1911.04944},
      archivePrefix={arXiv},
      primaryClass={cs.CL},
      url={https://arxiv.org/abs/1911.04944}, 
}

@misc{fan2020englishcentricmultilingualmachinetranslation,
      title={Beyond English-Centric Multilingual Machine Translation}, 
      author={Angela Fan and Shruti Bhosale and Holger Schwenk and Zhiyi Ma and Ahmed El-Kishky and Siddharth Goyal and Mandeep Baines and Onur Celebi and Guillaume Wenzek and Vishrav Chaudhary and Naman Goyal and Tom Birch and Vitaliy Liptchinsky and Sergey Edunov and Edouard Grave and Michael Auli and Armand Joulin},
      year={2020},
      eprint={2010.11125},
      archivePrefix={arXiv},
      primaryClass={cs.CL},
      url={https://arxiv.org/abs/2010.11125}, 
}

@misc{martins2025eurollm9btechnicalreport,
      title={EuroLLM-9B: Technical Report}, 
      author={Pedro Henrique Martins and João Alves and Patrick Fernandes and Nuno M. Guerreiro and Ricardo Rei and Amin Farajian and Mateusz Klimaszewski and Duarte M. Alves and José Pombal and Nicolas Boizard and Manuel Faysse and Pierre Colombo and François Yvon and Barry Haddow and José G. C. de Souza and Alexandra Birch and André F. T. Martins},
      year={2025},
      eprint={2506.04079},
      archivePrefix={arXiv},
      primaryClass={cs.CL},
      url={https://arxiv.org/abs/2506.04079}, 
}

@misc{oepen2026hplt30largescalemultilingual,
      title={HPLT 3.0: Very Large-Scale Multilingual Resources for LLMs and MT. Mono- and Bi-lingual Data, Multilingual Evaluation, and Pre-Trained Models}, 
      author={Stephan Oepen and Nikolay Arefev and Mikko Aulamo and Marta Bañón and Maja Buljan and Laurie Burchell and Lucas Charpentier and Pinzhen Chen and Mariya Fedorova and Ona de Gibert and Barry Haddow and Jan Hajič and Jindřich Helcl and Andrey Kutuzov and Veronika Laippala and Zihao Li and Risto Luukkonen and Bhavitvya Malik and Vladislav Mikhailov and Amanda Myntti and Dayyán O'Brien and Lucie Poláková and Sampo Pyysalo and Gema Ramírez Sánchez and Janine Siewert and Pavel Stepachev and Jörg Tiedemann and Teemu Vahtola and Dušan Variš and Fedor Vitiugin and Tea Vojtěchová and Jaume Zaragoza},
      year={2026},
      eprint={2511.01066},
      archivePrefix={arXiv},
      primaryClass={cs.CL},
      url={https://arxiv.org/abs/2511.01066}, 
}

@misc{feng2022languageagnosticbertsentenceembedding,
      title={Language-agnostic BERT Sentence Embedding}, 
      author={Fangxiaoyu Feng and Yinfei Yang and Daniel Cer and Naveen Arivazhagan and Wei Wang},
      year={2022},
      eprint={2007.01852},
      archivePrefix={arXiv},
      primaryClass={cs.CL},
      url={https://arxiv.org/abs/2007.01852}, 
}

@misc{dettmers2023qloraefficientfinetuningquantized,
      title={QLoRA: Efficient Finetuning of Quantized LLMs}, 
      author={Tim Dettmers and Artidoro Pagnoni and Ari Holtzman and Luke Zettlemoyer},
      year={2023},
      eprint={2305.14314},
      archivePrefix={arXiv},
      primaryClass={cs.LG},
      url={https://arxiv.org/abs/2305.14314}, 
}

@misc{post2018clarityreportingbleuscores,
      title={A Call for Clarity in Reporting BLEU Scores}, 
      author={Matt Post},
      year={2018},
      eprint={1804.08771},
      archivePrefix={arXiv},
      primaryClass={cs.CL},
      url={https://arxiv.org/abs/1804.08771}, 
}

@inproceedings{popovic-2015-chrf,
    title = "chr{F}: character n-gram {F}-score for automatic {MT} evaluation",
    author = "Popovi{\'c}, Maja",
    editor = "Bojar, Ond{\v{r}}ej  and
      Chatterjee, Rajan  and
      Federmann, Christian  and
      Haddow, Barry  and
      Hokamp, Chris  and
      Huck, Matthias  and
      Logacheva, Varvara  and
      Pecina, Pavel",
    booktitle = "Proceedings of the Tenth Workshop on Statistical Machine Translation",
    month = sep,
    year = "2015",
    address = "Lisbon, Portugal",
    publisher = "Association for Computational Linguistics",
    url = "https://aclanthology.org/W15-3049/",
    doi = "10.18653/v1/W15-3049",
    pages = "392--395"
}

@inproceedings{papineni-etal-2002-bleu,
    title = "{B}leu: a Method for Automatic Evaluation of Machine Translation",
    author = "Papineni, Kishore  and
      Roukos, Salim  and
      Ward, Todd  and
      Zhu, Wei-Jing",
    editor = "Isabelle, Pierre  and
      Charniak, Eugene  and
      Lin, Dekang",
    booktitle = "Proceedings of the 40th Annual Meeting of the Association for Computational Linguistics",
    month = jul,
    year = "2002",
    address = "Philadelphia, Pennsylvania, USA",
    publisher = "Association for Computational Linguistics",
    url = "https://aclanthology.org/P02-1040/",
    doi = "10.3115/1073083.1073135",
    pages = "311--318"
}

@inproceedings{rei-etal-2020-comet,
    title = "{COMET}: A Neural Framework for {MT} Evaluation",
    author = "Rei, Ricardo  and
      Stewart, Craig  and
      Farinha, Ana C  and
      Lavie, Alon",
    editor = "Webber, Bonnie  and
      Cohn, Trevor  and
      He, Yulan  and
      Liu, Yang",
    booktitle = "Proceedings of the 2020 Conference on Empirical Methods in Natural Language Processing (EMNLP)",
    month = nov,
    year = "2020",
    address = "Online",
    publisher = "Association for Computational Linguistics",
    url = "https://aclanthology.org/2020.emnlp-main.213/",
    doi = "10.18653/v1/2020.emnlp-main.213",
    pages = "2685--2702"
}

@misc{theomnilingualmtteam2025bouquetdatasetbenchmarkopen,
      title={BOUQuET: dataset, Benchmark and Open initiative for Universal Quality Evaluation in Translation}, 
      author={The Omnilingual MT Team and Pierre Andrews and Mikel Artetxe and Mariano Coria Meglioli and Marta R. Costa-jussà and Joe Chuang and David Dale and Cynthia Gao and Jean Maillard and Alex Mourachko and Christophe Ropers and Safiyyah Saleem and Eduardo Sánchez and Ioannis Tsiamas and Arina Turkatenko and Albert Ventayol-Boada and Shireen Yates},
      year={2025},
      eprint={2502.04314},
      archivePrefix={arXiv},
      primaryClass={cs.CL},
      url={https://arxiv.org/abs/2502.04314}, 
}

@misc{ociepa2025bielik11bv3multilingual,
      title={Bielik 11B v3: Multilingual Large Language Model for European Languages}, 
      author={Krzysztof Ociepa and Łukasz Flis and Remigiusz Kinas and Krzysztof Wróbel and Adrian Gwoździej},
      year={2025},
      eprint={2601.11579},
      archivePrefix={arXiv},
      primaryClass={cs.CL},
      url={https://arxiv.org/abs/2601.11579}, 
}

@misc{kocoń2025pllumfamilypolishlarge,
      title={PLLuM: A Family of Polish Large Language Models}, 
      author={Jan Kocoń and Maciej Piasecki and Arkadiusz Janz and Teddy Ferdinan and Łukasz Radliński and Bartłomiej Koptyra and Marcin Oleksy and Stanisław Woźniak and Paweł Walkowiak and Konrad Wojtasik and Julia Moska and Tomasz Naskręt and Bartosz Walkowiak and Mateusz Gniewkowski and Kamil Szyc and Dawid Motyka and Dawid Banach and Jonatan Dalasiński and Ewa Rudnicka and Bartłomiej Alberski and Tomasz Walkowiak and Aleksander Szczęsny and Maciej Markiewicz and Tomasz Bernaś and Hubert Mazur and Kamil Żyta and Mateusz Tykierko and Grzegorz Chodak and Tomasz Kajdanowicz and Przemysław Kazienko and Agnieszka Karlińska and Karolina Seweryn and Anna Kołos and Maciej Chrabąszcz and Katarzyna Lorenc and Aleksandra Krasnodębska and Artur Wilczek and Katarzyna Dziewulska and Paula Betscher and Zofia Cieślińska and Katarzyna Kowol and Daria Mikoś and Maciej Trzciński and Dawid Krutul and Marek Kozłowski and Sławomir Dadas and Rafał Poświata and Michał Perełkiewicz and Małgorzata Grębowiec and Maciej Kazuła and Marcin Białas and Roman Roszko and Danuta Roszko and Jurgita Vaičenonienė and Andrius Utka and Paweł Levchuk and Paweł Kowalski and Irena Prawdzic-Jankowska and Maciej Ogrodniczuk and Monika Borys and Anna Bulińska and Wiktoria Gumienna and Witold Kieraś and Dorota Komosińska and Katarzyna Krasnowska-Kieraś and Łukasz Kobyliński and Martyna Lewandowska and Marek Łaziński and Mikołaj Łątkowski and Dawid Mastalerz and Beata Milewicz and Agnieszka Anna Mykowiecka and Angelika Peljak-Łapińska and Sandra Penno and Zuzanna Przybysz and Michał Rudolf and Piotr Rybak and Karolina Saputa and Aleksandra Tomaszewska and Aleksander Wawer and Marcin Woliński and Joanna Wołoszyn and Alina Wróblewska and Bartosz Żuk and Filip Żarnecki and Konrad Kaczyński and Anna Cichosz and Zuzanna Deckert and Monika Garnys and Izabela Grabarczyk and Wojciech Janowski and Sylwia Karasińska and Aleksandra Kujawiak and Piotr Misztela and Maria Szymańska and Karolina Walkusz and Igor Siek and Jakub Kwiatkowski and Piotr Pęzik},
      year={2025},
      eprint={2511.03823},
      archivePrefix={arXiv},
      primaryClass={cs.CL},
      url={https://arxiv.org/abs/2511.03823}, 
}

@misc{kudugunta2023madlad400,
      title={MADLAD-400: A Multilingual And Document-Level Large Audited Dataset}, 
      author={Sneha Kudugunta and Isaac Caswell and Biao Zhang and Xavier Garcia and Christopher A. Choquette-Choo and Katherine Lee and Derrick Xin and Aditya Kusupati and Romi Stella and Ankur Bapna and Orhan Firat},
      year={2023},
      eprint={2309.04662},
      archivePrefix={arXiv},
      primaryClass={cs.CL}
}

@article{Siuciak_2025,
      title="{„Język śląski” – czyli o mieszaniu porządków}",
      volume={32},
      url={https://pressto.amu.edu.pl/index.php/pspsj/article/view/51890},
      DOI={10.14746/pspsj.2025.32.2.24},
      number={2},
      journal={Poznańskie Studia Polonistyczne. Seria Językoznawcza},
      author={Siuciak, Mirosława},
      year={2025},
      month={grudz.},
      pages={329–346}
}

@article{dubisz_2024,
      title={Lingwistycznie o godce ślunskij i języku regionalnym},
      volume={815},
      DOI={10.33896/PorJ.2024.6.9},
      number={6},
      journal={Poradnik Językowy},
      publisher={Towarzystwo Kultury Języka},
      author={Dubisz, Stanisław},
      year={2024},
      pages={105–110}
}

@article{hentschel_2018,
      title={Śląski: Gwara – dialekt – język? Spojrzenie z zewnątrz},
      volume={4},
      url={https://sbc.org.pl/dlibra/publication/453528},
      journal={Polonistyka na początku XXI wieku : diagnozy, koncepcje, perspektywy. T. 4 : Pogranicza, mniejszości, regiony, etnolingwistyka},
      publisher={Wydawnictwo Uniwersytetu Śląskiego},
      language={pol},
      author={Hentschel, Gerd},
      year={2018},
      pages={41–66}
}

@article{bandtkie,
    author = {Bandtkie, Jerzy Samuel},
    title = {Wiadomości o ięzyku Polskim w Szląsku i o Polskich Szlązakach},
    journal = {Mrówka Poznańska},
    year = 1821,
    publisher={J. A. Munk},
    language={pol},
    title={Mrówka Poznańska : pismo ku użyteczney zabawie rozumu i serca 1821.03 R.1 T.1},
}

@book{steuer,
    author = {Steuer, Feliks},
    title = {Ostatńi gwojźdźaurz},
    publisher = {Księgarnia i Drukarnia Katolicka S. A. Katowice},
    year = {1935}
}

@misc{polgov2020,
  author = "{Government of Poland}",
  title = "{Comments of the Government of Poland on the Fourth Opinion of the Advisory Committee on the implementation of the Framework Convention for the Protection of National Minorities by Poland}",
  address = {Warsaw},
  howpublished = {official memorandum},
  year = {2020},
  url = {https://rm.coe.int/4th-com-poland-en/16809e4dbf}
}

@unpublished{przywara,
    author = {Przywara, Michał},
    title = {Narzecza śląskie},
    note = {unpublished manuscript},
    year = {before 1904},
    url = {https://www.sbc.org.pl/dlibra/publication/68791/edition/64871}
}

@book{olesch,
 author={Olesch, Reinhold},
 publisher={Osteuropa-Institut},
 language={ger},
 title={Der Wortschatz der polnischen Mundart von Sankt Annaberg},
 volume={1}
}

@article{zareba,
    author = {Zaręba, Alfred},
    title = "{Słownik Starych Siołkowic w powiecie opolskim}",
    journal = {Zeszyty naukowe Uniwersytetu Jagiellońskiego / Rozprawy i studia},
    year = {1960},
    publisher = {Nakł. Uniwersytetu Jagiellońskiego}
}

@article{pluta,
    author = {Pluta, Feliks},
    title = "{Słownictwo Dzierżysławic w powiecie prudnickim}",
    journal = {Zeszyty naukowe Wyższej Szkoły Pedagogicznej im. Powstańców Śląskich w Opolu / Seria B: Studia i monografie},
    year = {1973},
    publisher = {Państwowe Wydawnictwo Naukowe – Oddział Wrocławski}
}

@book{czastkaszymon,
    author = {Cząstka-Szymon, Bożena and Ludwig, Jerzy and Synowiec, Helena},
    title = {Mały słownik gwary Górnego Śląska},
    year = {2000},
    publisher = {Leksykon}
}

@phdthesis{nikel,
    author = {Nikel, Emmanuel},
    title = {Die polnische Mundart des oberschlesischen Industriebezirks},
    school = {Friedrich-Wilhelms-Universität zu Berlin},
    year = {1908} 
}

@book{kaisig,
    author = {Spectator},
    title = {Sprachmischung in Oberschlesien. Ein Beitrag zur oberslesischen Volkskunde},
    year = {1908},
    publisher = {Verlag von Gebrüder Böhm}
}

@article{koszyk,
    author = {Koszyk, Szymon},
    title = {Z gwary „wojouków” opolskich},
    journal = {Kwartalnik opolski},
    year = {1957},
    volume = {2},
    pages = {168-184}
}

@book{stawoj,
    author = {Iżykowska, Małgorzata and Starczewska–Wojnar, Aleksandra},
    title = {Przeczytano, przyjęto, podpisano: polskie protokoły sądów rozjemczych w powiecie strzeleckim w XIX wieku: edycja źródłowa ze zbiorów Archiwum Państwowego w Opolu},
    year = {2017},
    publisher = {Archiwum Państwowe w Opolu},
    url = {https://repo.uni.opole.pl/info/book/UO329b22b0ccd44eed8437e0b8cf59805a/}
}

@report{gus2021,
  author       = {{Statistics Poland}},
  title        = {Preliminary results of the National Population and Housing Census 2021 in the scope of national-ethnic structure and language used in home contacts},
  year         = {2022},
  institution  = {Statistics Poland (G{\l}ówny Urz{\k{a}}d Statystyczny)},
  url          = {https://stat.gov.pl/download/gfx/portalinformacyjny/pl/defaultaktualnosci/6494/10/1/1/wstepne_wyniki_nsp_2021_w_zakresie_struktury_narodowo-etnicznej_oraz_jezyka_kontaktow_domowych.pdf},
  note         = {Accessed: 2026-05-25}
}

@book{malinowski,
    author = {Malinowski, Lucjan},
    title = {Powieści ludu polskiego na Śląsku},
    year = {1899},
    publisher = {Akademia Umiejętności}
}

\appendix

\section{Silesian Language Description}
\label{app:silesian_language}

Despite the efforts of documenting the Silesian language reaching the first half of the 19th century~\cite{bandtkie}, the proposals for a single orthography had not reached a wider consensus until 2009 when the \textit{ślabikŏrz} (meaning: ``primer'') orthography was agreed. Until then the few Silesian books and other publications used an ad hoc modified Polish orthography or -- as in the case of a linguist Feliks Steuer’s works -- a special phonetic orthography~\cite{steuer}.

As a signatory of the European Charter for Regional or Minority Languages, Poland recognises languages of its national and ethnic minorities as well as Kashubian as a regional language. There have been several attempts to recognise Silesian as the second regional language, however all of them failed. The opponents of this solution argue that the term \textit{regional language} does not cover dialects of the official state language \cite[see eg.][page 10]{polgov2020}. This led to an intense discourse about the classification of Silesian as a separate language and not a dialect of Polish.

As proposed by~\citet{hentschel_2018}, Silesian could be described as Abstand language or Abstand lect. Silesian has isolated itself from neighbouring dialects and Polish standard language. Since Polish dialects around Silesia have died out, there is no basis to consider Silesian a part of Polish dialect continuum.

On the other hand the supporters of preserving the status quo argue that Silesian is mutually intelligible with the Polish standard language and that the features of the Silesian language can be also found in other Polish dialects \cite{dubisz_2024,Siuciak_2025}.

The lack of recognition barres the language from being taught at schools. Because of that a relatively low number of Silesians have mastered the new orthography. Although Silesian cultural organisations use it in their public communication, most of Silesian speakers still tend to phonetically write using Polish orthography. This is further reinforced by a caveat that could be named a sociotechnical one: there are no electronical devices offering Silesian keyboard by default or even as a preinstalled option. Silesian speakers have to deliberately download and install it. This usually leads to resorting to the Polish keyboard.

Silesian exhibits variation across orthography (non-standardized spelling conventions), regional morphology, and lexical borrowings from neighbouring languages.
These properties make direct sentence-level translation challenging for purely data-driven systems, especially in low-resource regimes.

\section{Rule-based system pipeline}
\label{app:pipeline}

To illustrate the rule-based system pipeline, consider the phrase:

\textit{system przekładający maszynowo}

“machine-translating system”

The system first analyses the phrase using Constraint Grammar. The first of the three tokens “system” can have two morphological intepretations, the second “przekładający” can have nine, and finally “maszynowo” can have two. Constraint Grammar eliminates these ambiguities and leaves the three correct ones:

\texttt{"system" n mi sg nom}

\texttt{"przekładać" vblex impf pprs mi sg nom}

\texttt{"maszynowo" adv}

The disambiguated representation is then sent to the lexical transfer module which translates the result using the bilingual dictionary:

\textit{systym przekłŏdajōncy maszinowo}

“machine-translating system”

The phrase would be understandable to a Silesian speaker, however this is a seldom used form. The structural transfer module detects the pattern and applies a rule for a more natural sounding output:

noun + present participle used adjectivally → noun + comma + “co” + present participle retagged as third-person present

The result is:

\textit{systym, co przekłŏdŏ maszinowo}

“system that machine-translates”

Despite its simplicity, the system provides strong performance in low-resource settings due to explicit linguistic priors.

\section{Neural models}
\label{sec:neural_models}

\subsection{NLLB: No Language Left Behind}
\textbf{NLLB-200-3.3B} and \textbf{NLLB 54B MOE} this MT models are included due to its explicit support for Silesian in its training data.

\subsection{GPT-5.4}
We include this model as state-of-the-art commercial benchmark. While its scale differs significantly from our models, it provides a vital reference for how frontier LLM handle low-resource tasks in a zero-shot setting. We obtained these translations through the OpenAI API.


The model was evaluated with default settings, without reasoning enabled, and with temperature set to 1.0. The prompt used for translation is shown in Figure~\ref{fig:silesian-prompt}. The prompt was written in Polish, as it was the source language of the translation task and provided the most natural interaction setting for the evaluated model.
\begin{figure}[h]
\centering
\begin{quote}
\texttt{Przetłumacz poniższe zdanie na język śląski. \\
Zwróć wyłącznie tłumaczenie po śląsku
(bez komentarzy, wyjaśnień ani dodatkowego tekstu). \\[0.5em]
Tekst: \{text\}}
\end{quote}
\caption{Prompt used for Silesian translation with GPT-5.4, Bielik and PLLum models. English translation: ``Translate the following sentence into the Silesian language. Return only the translation in Silesian (without comments, explanations, or additional text). Text: \{text\}.''}
\label{fig:silesian-prompt}
\end{figure}

\subsection{Google Translate}
This system is included as a representative industry standard for production-grade translation quality. To ensure reproducibility, we note that the translations were processed in April and May 2026, through the Google Cloud Translation API~\cite{wu2016googlesneuralmachinetranslation},

\subsection{TranslateGemma-4B-IT (Base and Fine-tuned)}
We evaluate both base and custom fine-tuned variants, the details of which are provided in section~\ref{sec:experiments}. ~ \cite{finkelstein2026translategemmatechnicalreport}

We selected this model because it was trained on the English--Silesian translation direction, which we expected could transfer well to the Polish--Silesian translation task. However, the model's chat template did not include support for the Silesian language. To address this limitation, we manually extended the chat template by adding the Silesian language support as "szl": "Silesian (Śląski)" and then performed inference using the modified chat template.

\subsection{MADLAD-400-10B-MT}

MADLAD-400-10B-MT is a multilingual machine translation model based on the T5 architecture. It was trained on 250 billion tokens of publicly available data covering more than 450 languages, including Silesian.

\subsection{Bielik-PL-11B-v3.0-IT}

This is a multilingual model with a strong emphasis on Polish. Notably, it was also pretrained on Silesian-language data. We used the system prompt shown in Figure~\ref{fig:silesian-prompt_system_bielik} and the user prompt shown in Figure~\ref{fig:silesian-prompt}.
\begin{figure}[h]
\centering
\begin{quote}
\texttt{Jesteś ekspertem od tłumaczeń polsko-śląskich.}
\end{quote}
\caption{System prompt used for Silesian translation with Bielik model. English translation: ``You are an expert in Polish-Silesian translations.''}
\label{fig:silesian-prompt_system_bielik}
\end{figure}

\subsection{PLLuM-12B-nc-chat}

This model was optimized for Polish and designed for general-purpose chat, question answering, and text generation tasks. We used the user prompt shown in Figure~\ref{fig:silesian-prompt}.

\subsection{Prompt Configuration}

To ensure a fair comparison across models, we kept the prompting configuration as close as possible to the default settings provided by each model interface. The goal of this study was to compare the translation capabilities of the evaluated models rather than optimize their performance through extensive prompt engineering or parameter tuning. For this reason, we avoided additional system-level modifications, handcrafted instructions, or decoding adjustments beyond the prompts presented in Figures~\ref{fig:silesian-prompt} and~\ref{fig:silesian-prompt_system_bielik}.

\section{Expanded results}
\label{app:expanded_results}

\subsection{Silesian to Polish direction}
\label{sec:szl2pl}

Although our primary focus is the Polish-to-Silesian direction, we additionally evaluate on the reverse Silesian-to-Polish translation task using the BOUQuET benchmark. Since Polish is a high-resource language, this direction is generally easier for translation systems and generative models.

Table~\ref{table:bouquet_szl_pl} presents the evaluation results. GPT-5.4 achieves the best overall performance with a BLEU score of 41.1 and chrF of 63.2, slightly outperforming Google Translate. The rule-based system also achieves competitive results, reaching 31.7 BLEU and 57.4 chrF, outperforming both TranslateGemma and NLLB 3.3B. This suggests that manually designed linguistic rules and dictionaries remain effective for closely related language pairs such as Silesian and Polish, where systematic lexical and orthographic correspondences can be captured explicitly. TranslateGemma obtains moderate results, while NLLB 3.3B performs substantially worse than the strongest proprietary systems. In many cases, NLLB 3.3B failed to translate Silesian-specific words and expressions, leaving them unchanged in the Polish output.

\begin{table}[h]
\centering
\small
\begin{tabular}{lrr}
\toprule
\textbf{Model} & \textbf{BLEU} & \textbf{chrF} \\
\midrule
NLLB 3.3B & 18.0 & 32.5 \\
TransGemma & 24.8 & 48.4 \\ \midrule
GPT-5.4 & \textbf{41.1} & \textbf{63.2} \\
Google Translate & 38.3 & 61.7 \\ \midrule
\small{\textit{ours}} \\
Rule-Based & 31.7 & 57.4 \\
\bottomrule
\end{tabular}
\caption{Evaluation results on the BOUQuET SZL$\rightarrow$POL translation task.}
\label{table:bouquet_szl_pl}
\end{table}

\end{document}